\documentclass[journal]{IEEEtran}
\ifCLASSINFOpdf
\else
   \usepackage[dvips]{graphicx}
\fi
\usepackage{url}
\usepackage[T1]{fontenc}
\usepackage{graphicx}
\usepackage{amsmath,amssymb,amsfonts}
\usepackage{algorithmic}
\usepackage{xcolor}
\usepackage{multirow}
\usepackage{color}
\usepackage{float}

\usepackage{balance}

\begin{document}

\title{Multi-modal Visual Place Recognition in Dynamics-Invariant Perception Space}

\author{ Lin  Wu, Teng Wang, and Changyin Sun {	
\thanks{This work was supported by in part by Postgraduate Research \& Practice Innovation Program of Jiangsu Province under Grant SJCX20\_0035, and in part by Fundamental Research Funds for the Central Universities under Grant 3208002102D, and in part by National Natural Science Foundation of China under Grant 61803084. (Corresponding author: Teng Wang)}
}
\thanks{L. Wu, T. Wang and C. Sun are with School of Automation, Key Laboratory of Measurement and Control of Complex System of Engineering, Southeast University, Nanjing 210096, China (e-mail: wulin@seu.edu.cn; wangteng@seu.edu.cn; cysun@seu.edu.cn)}
}
\markboth{Journal of \LaTeX\ Class Files}
{Shell \MakeLowercase{\textit{et al.}}: Bare Demo of IEEEtran.cls for IEEE Journals}
\maketitle

\begin{abstract}

Visual place recognition is one of the essential and challenging problems in the fields of robotics. In this letter, we for the first time explore the use of multi-modal fusion of semantic and visual modalities in dynamics-invariant space to improve place recognition in dynamic environments. We achieve this by first designing a novel deep learning architecture to generate the static semantic segmentation and recover the static image directly from the corresponding dynamic image. We then innovatively leverage the spatial-pyramid-matching model to encode the static semantic segmentation into feature vectors. In parallel, the static image is encoded using the popular Bag-of-words model. On the basis of the above multi-modal features, we finally measure the similarity between the query image and target landmark by the joint similarity of their semantic and visual codes. Extensive experiments demonstrate the effectiveness and robustness of the proposed approach for place recognition in dynamic environments.

\end{abstract}


\IEEEpeerreviewmaketitle

\section{Introduction}
\IEEEPARstart{V}{isual} place recognition(VPR), as the key component of a SLAM system~\cite{wald2020,2020Robust}, is a task to help a robot determine whether it locates in a previously visited place. Current works typically regard it as an image retrieval task which matches the current observation against a set of reference landmarks~\cite{Ding_2019_ICCV, wald2020}, and design various feature descriptors to measure the landmark similarity~\cite{Mur2017ORB,Engel2016Direct,2017ORB}. These methods are conceived to operate in static environments. However, the real world is complex and dynamic. The presences of dynamic objects make the scene appearance not consistent at different moments, thus increasing the errors of feature matching. Therefore, it is of great importance to improve the robustness of feature matching in dynamic environments.

At present, one popular solution to handle dynamic scenes is to detect moving objects in the scene and eliminate their negative influences on feature matching by discarding them as outliers. For example, Scona et al.~\cite{2018StaticFusion} estimate a probabilistic static/dynamic segmentation of the input RGB image and use the segmentation to weight each pixel. Xiao et al.~\cite{2019Dynamic} employ SSD~\cite{2016SSD} to detect dynamic objects and excludes dynamic feature points from feature matching. An alternative method is to translate the dynamic images into realistic static frames, and perform feature matching on recovered static images. The most related work is by Berta et al.~\cite{bescos2019empty}, who improve Pix2Pix~\cite{isola2017image} by performing conditioning on both the dynamic image and its dynamic mask under cGAN~\cite{mirza2014conditional}, to recover realistic static images. Recently, Berta et al.~\cite{2020Empty} implement two more losses based on image steganalysis techniques and ORB features, respectively, to better recover reliable features. 

Although these mainstream approaches enhance feature matching in dynamic environments, they have their own drawbacks. On one hand, simply discarding dynamic contents as done in \cite{2018StaticFusion,2019Dynamic} reduces the amount of available features and may cause failures in feature matching when the dynamic portions of the image tends to dominate the whole image in feature space. On the other hand, although those dynamic-to-static image translation approaches~\cite{bescos2019empty,2020Empty} are capable of generating visually realistic images, they easily introduce blur and artifacts, especially in areas associated with moving objects. Extracting features on such recovered static images will degrade feature matching to some extent.

In this work, we share a similar line of thought with Berta et al.~\cite{bescos2019empty, 2020Empty}, but move one step forward to build a multi-modal dynamics-invariant perception space to improve feature matching in dynamic environments. This space is built by first designing a novel deep neural network architecture to reconstruct the static semantics  (i.e., static semantic segmentation map) and static images from the dynamic images in a sequential manner. Here, the recovered static semantics are treated as priors to guide the generation of the static images with high semantic consistency. To the best of our knowledge, we are the first to study the reconstruction of static semantics directly from dynamic images. We then encode the recovered static semantics and static images into feature descriptors using spatial-pyramid-matching (SPM) and Bag-of-words (BOW) models, respectively. This multi-modal features are finally utilized to jointly determine the image similarity. Main contributions of this letter are summarized as follows: (1) We propose a dynamics-invariant perception network to recover the static semantics and static images directly from the dynamic frames; (2) We design a multi-modal coding strategy to generate the robust semantic-visual features for image matching; (3) The superiority of the proposal is verified through extensive experiments.

\begin{figure*}
	\centering
	\includegraphics[height=0.21\textheight,width=\linewidth]{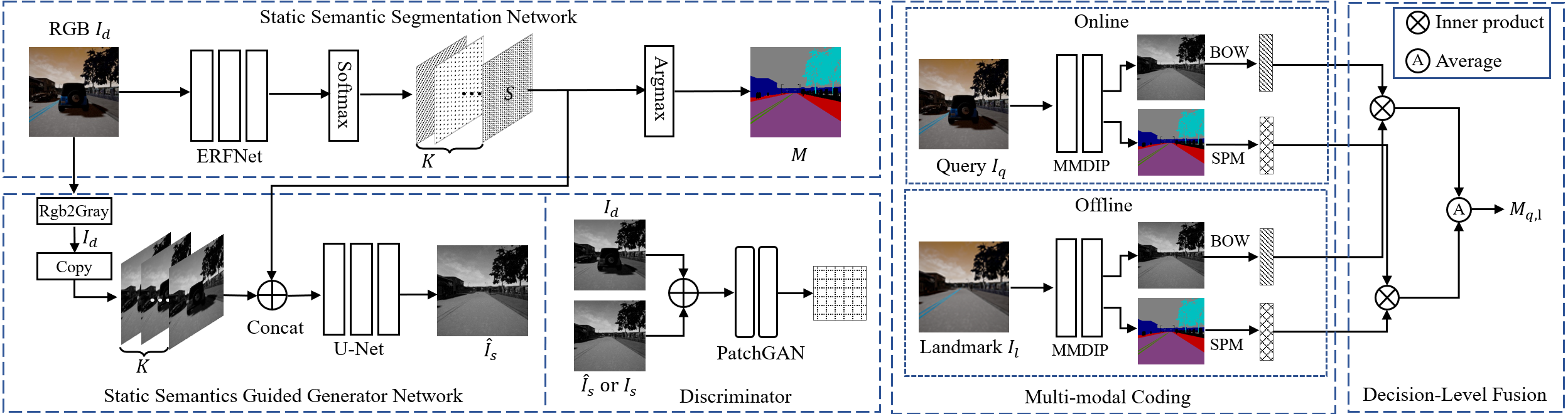}
	\caption{The overall framework of our proposed multi-modal VPR in dynamics-invariant perception space.}
	\label{fig_framework}
\end{figure*}

\section{Proposed Approach}
\subsection{Overall Framework}
The overall framework of our multi-modal VPR approach is shown in Fig.~\ref{fig_framework}. To handle the presence of moving objects, we first reconstruct the static semantics and the static frame directly from the current query image using the multi-modal  dynamics-invariant perception (MMDIP) network, aiming to bridge the appearance inconsistency of the scene at different moments. Then, the static image and static semantics are embedded into different feature spaces by the multi-modal coding strategy. We finally measure the similarity between the query image and each reference landmark pre-stored in memory using the joint similarity of different modalities.

\subsection{Multi-modal Dynamics-invariant Perception}
\label{sub:divp}
\subsubsection{Static Semantic Segmentation} 
\label{subsub:ssm}
We regard this problem  as a novel semantic segmentation task and borrow recent ideas from related works to solve it. Because real-world applications pose a high requirement on both accuracy and speed~\cite{2021EACNet}, we prefer the Efficient Residual Factorized Network (ERFNet)~\cite{2017ERFNet}, which uses residual connections and factorized convolutions to achieve fast and accurate pixel-wise semantic segmentation. The final output of ERFNet  is a probability map $S \in \mathbb{R}^{H \times W \times K}$, with $S_{i, j, k}$ indicating probability of pixel $p_{i, j}$ at row $i$ and column $j$ that belongs to semantic class $k$. The network is trained using the weighted multi-class cross entropy loss~\cite{yu2018bisenet}:
\begin{equation}\label{f1}
\mathcal{L}_{seg}=-\frac{1}{\log (r_c+1+\epsilon)} \cdot \log (S_{i,j,c})
\end{equation}
where $c$ denotes the ground-truth label of pixel $p_{i,j}$, $r_{c}$ refers to the proportion of the $c$-th category in the training set, and $\epsilon$ is set to 0.02. The loss is summed up over all the pixels in a mini-batch. After training procedure, the static semantic segmentation map $M$ is obtained by assigning each pixel to the category with maximum probability. 

\subsubsection{Dynamic-to-Static Image Translation} We explore cGAN consisting of a generator $G$ and a discriminator $D$ to recover static images. $G$ uses U-Net~\cite{ronneberger2015u} to learn a mapping from observed dynamic image $I_d$ and estimated semantic probability map $S$ to static image $\hat{I}_{s}$, $G: (I_d, S)\rightarrow \hat{I}_{s}$. Here, the grayscale dynamic image $I_d$ is first copied $K$ times and then concatenated to $S$ for alleviating channel imbalance.  It is worth noting that we prefer the semantic probability map $S$ over the semantic segmentation map $M$ as inputs in order to handle noises from ERFNet while guiding $G$ to produce semantic-consistent static images.  Adversarially, $D$ receives synthetic static image $\hat{I}_{s}$ or real static image $I_s$ concatenated with its corresponding dynamic image $I_d$ as inputs and use PatchGAN discriminator~\cite{isola2017image} to produce a probability map, with each element indicating the likelihood of corresponding patch in the input image $\hat{I}_{s}$  or $I_s$ being real. 

The image translation network is trained using standard losses. For the generator, we apply adversarial loss~\cite{mao2017least}, pixel loss~\cite{isola2017image}, and perceptual loss~\cite{johnson2016perceptual}. For the discriminator, only adversarial loss~\cite{mao2017least} is used to distinguish whether the input image is real or fake. The generator and discriminator are trained alternately to play a min-max game. The generator loss $\mathcal{L}_G$ and the discriminator loss $\mathcal{L}_D$ can be written as:
\vspace{0.1cm}
\begin{equation}\label{gloss}
  \mathcal{L}_{G}=\mathbb{E}[\parallel D(\hat{I}_s \oplus I_d)-1\parallel_2] + \mathcal{L}_{PI} (\hat{I}_s, I_s) + \mathcal{L}_{PE} (\hat{I}_s, I_s) 
\end{equation}
\begin{equation}\label{dloss}
  \mathcal{L}_{D}=\mathbb{E}[\parallel D(I_s \oplus I_d)-1\parallel_2] + \mathbb{E}[\parallel D(\hat{I}_s \oplus I_d)\parallel_2]
\end{equation}
where  $\oplus$ denote concatenation along channels; $\mathcal{L}_{PI}$ is the pixel loss that measures the L1 distance between the produced static image $\hat{I}_s$ and target image $I_s$; $\mathcal{L}_{PE}$ is the perceptual loss that measures the differences between the feature representations extracted by VGG-16. In the following, we refer to the static-semantics-guided generator network as SSGGNet. 





\subsection{Multi-modal Coding}
\label{sub:mmvr}
\subsubsection{BoW-based Image Coding} Currently, BoW approach is one of the most popular methods to measure image similarity for identifying previously visited places in various SLAM systems~\cite{Mur2017ORB,Engel2016Direct,2017ORB}.  Main idea of BoW is to quantize local descriptors into visual words and then represent each image as a vector of words~\cite{2018Saliency, Sch2018}. In this work, we follow previous works to rely on BoW model to encode each recovered static image into a feature vector $g$. 

\subsubsection{SPM-based Semantics Coding} BoW model represents an image as a disordered set of local descriptors. It ignores the spatial layout  of features, and thus has severely limited representational capacity. We alleviate this issue by introducing position-sensitive semantic features, since layout information is quite important to differentiate segmentation maps. We draw idea from spatial pyramid matching (SPM) model~\cite{lazebnik2006}, and develop a novel semantics coding strategy. First, construct a sequence of increasingly finer grids at resolution $0,\cdots, L$ on the estimated static semantic segmentation map $M$, such that the grid at level $l$ has $2^l\times 2^l$ cells. Then, count the number of pixels belonging to each semantic category in each grid cell to form a descriptor of length $2^{2l}\times K$ for each level $l$, $h^l=[h_1^l;h_2^l;...;h_{2^{2l}}^l]$. At last, assign different weights to the descriptors at different levels, and concatenate them to form the multi-scale feature descriptors $h=[\hat{h}^0;\hat{h}^1;...;\hat{h}^L]$. Here, $\hat{h}^l$ is the weighted feature descriptor $\hat{h}^l=w^{l}\cdot h^l$, with weight $w^l$ defined as follows.

\begin{equation}
	w^{l}=
	\begin{cases}
	2^{-L} & l=0\\
	2^{l-L-1} & l\neq0
	\end{cases}
\end{equation}

\subsection{Decision-level Multi-Modal Fusion} Given a pair of query image $I_q$ and reference landmark image $I_l$, we measure the cosine similarity of each modality separately, and utilize the weighted sum to measure the similarity between two input images as follows.
\begin{equation}\label{f11}
M_{q,l}=  \alpha <\!\frac{g_q}{||g_q||}, \frac{g_l}{||g_l||}\!> + (1\!-\!\alpha)
<\!\frac{h_q}{||h_q||}, \frac{h_l}{||h_l||}>\!
\end{equation}
where $(g_q, h_q)$ and $(g_l, h_l)$ are encoded BoW and SPM features from $I_q$ and $I_l$, respectively; $\alpha$ is a trade-off parameter to balance the importance of static images and static semantics, and is set to 0.5 by default.

\section{Experimental Results}

\subsection{Experimental Settings}
\label{sub:settings}
\textbf{Dataset.} Our dataset is built by CARLA, which is an open source simulator for autonomous driving research. CARLA not only provides massive digital resources, but also supports flexible configuration of environmental conditions and sensor suites. By controlling the presences of moving pedestrians and vehicles, we generate a total of 12,691 static-dynamic image pairs, with each pair captured at the same pose with the same illumination. In particular, different image pairs are captured from different locations, and each location is treated as one landmark for our VPR task. The whole dataset is split into training and test set, with 10,000 pairs randomly selected for training and the left 2,691 pairs for testing purpose.

\textbf{Implementation details.} We resize all images to $512\times 512$ before feeding them into our dynamics-invariant perception network. The two-phase training scheme is employed to train the whole perception model. In the first phase, the static semantic segmentation model is trained independently with Adam optimizer until convergence. The initial learning rate and batch size are set to $2\times10^{-3}$ and 1, respectively. In the second phase, we re-use the weight of pretrianed model to initialize the semantic segmentation network, and train the whole model in an end-to-end manner using Adam optimizer. We employ different initial learning rates of $1\times10^{-4}$, $2\times10^{-3}$ and $4\times10^{-3}$ for segmentation network, generative network and discriminative network, respectively, and set batch size to 4. The smaller learning rate is utilized to fine-tune the segmentation network, and the larger one is used to encourage efficient training of the discriminative network. Our model is implemented on PyTorch v1.2, CUDA v10.0, and run on Ubuntu 16.04 system with Intel (R) Xeon e5-2640 v4 CPU (2.4GHz) and NVIDIA TITAN Xp GPU (12 G memory).

\textbf{Evaluation metrics:} We follow previous works~\cite{2017ERFNet, 2021Nonlocal} to employ PA, MPA, MIoU and FWIoU to evaluate the performance of our static semantic segmentation model. As for image translation, we adopt the four popular metrics of L1, L2, PSNR and SSIM to evaluate the quality of the generated static images. Similar to ~\cite{liu2019lending, shi2020optimal}, we consider the recall accuracy at top-1, top-5, top-10, up to top-1\% (i.e., R@1, R@5, R@10, R@1\%) to evaluate the performance of our VPR method.

\subsection{Evaluations on Dynamics-invariant Perception}
\label{sub:semantic}
\begin{figure}[htbp]
	\centering	
	\begin{minipage}{\linewidth}
		\centerline{\includegraphics[width=\textwidth]{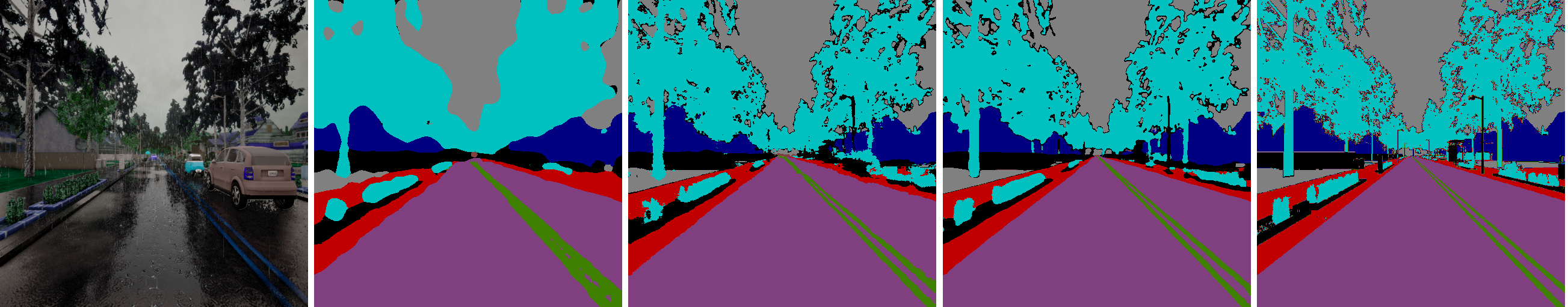}}
		\vspace{1pt}
	\end{minipage}
	\begin{minipage}{\linewidth}
		\centerline{\includegraphics[width=\textwidth]{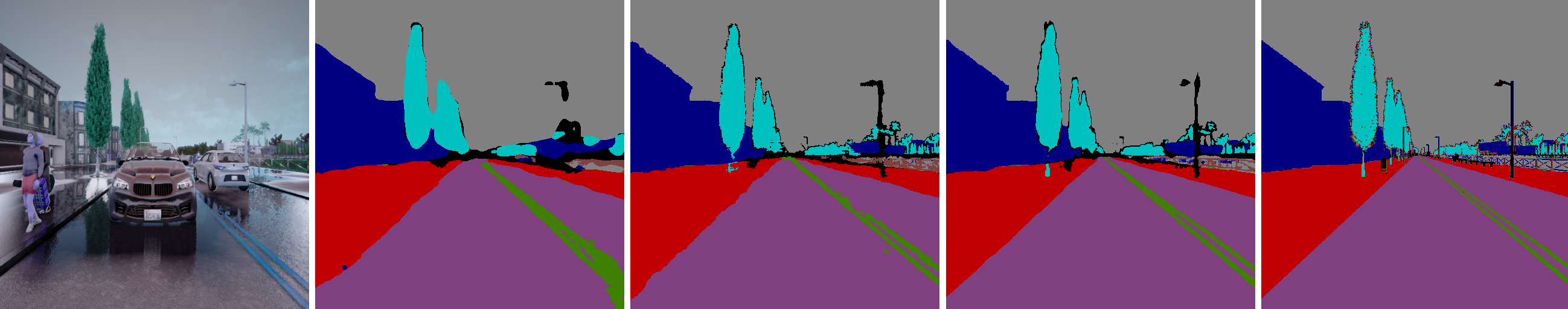}}
		\vspace{2pt}
	\end{minipage}
	\begin{minipage}{0.19\linewidth}
		\centerline{\footnotesize{(a) Dynamic}}
	\end{minipage}
	\begin{minipage}{0.19\linewidth}
		\centerline{\footnotesize{(b) FCN}}
	\end{minipage}
	\begin{minipage}{0.17\linewidth}
		\centerline{\footnotesize{(c) U-Net}}
	\end{minipage}
	\begin{minipage}{0.19\linewidth}
		\centerline{\footnotesize{(d) ERFNet}}
	\end{minipage}
	\begin{minipage}{0.19\linewidth}
		\centerline{\footnotesize{(e) GT}}
	\end{minipage}
	\caption{Visual evaluations on dynamics-invariant semantic segmentation.}
	\label{fig_seg}
\end{figure}

\textbf{Semantic Segmentation.} We explore several classic semantic segmentation networks for dynamic-to-static semantic segmentation, including FCN~\cite{2015Fully}, U-Net~\cite{ronneberger2015u} and ERFNet~\cite{2017ERFNet}. Numerical and visual evaluations of these approaches are reported in Table \ref{tab_seg} and Fig.~\ref{fig_seg}, respectively. We observe from the table that although all three models achieve impressive performances, U-Net and ERFNet outperform FCN by a large margin over all four metrics. We also notice from Fig.~\ref{fig_seg} that U-Net and ERFNet are capable of generating detailed and coherent static segmentation maps even in complex scenes with large dynamic portions. Despite the comparable performance from U-Net,  ERFNet shows much lower model complexity. We thus choose ERFNet for static semantic segmentation.

\vspace{-0.2cm}

\begin{table*}[htb]
	\caption{Numerical evaluations on semantic segmentation. ERFNet ranks first in all indicators of PA, MPA, MIoU, FWIoU.}
	\label{tab_seg}
	\centering
	\resizebox{\textwidth}{!}{
		\begin{tabular}{lc|llllllll|llll}
			\hline
			\multirow{2}{*}{Method} &\multirow{2}{*}{Params}& \multirow{2}{*}{None} & \multirow{2}{*}{Buildings} & \multirow{2}{*}{Fences} & \multirow{2}{*}{Others} & \multirow{2}{*}{Roadlines} & \multirow{2}{*}{Road} & \multirow{2}{*}{Sidewalk} & \multirow{2}{*}{Vegetation} & \multirow{2}{*}{PA} & \multirow{2}{*}{MPA} & \multirow{2}{*}{MIoU}& \multirow{2}{*}{FWIoU} \\
			&&&&&&&&&&&& \\
			\hline
			\multirow{1}{*}{FCN} 
			&134.5M &0.9342    &0.6316   &0.1866  &0.3455  &0.4316 &0.9492  &0.8283 &0.6978 &0.9223 &0.7441 &0.6256 &0.8771\\
			\hline
			\multirow{1}{*}{UNet} 
			&13.6M &0.9668    &0.6859   &0.2891  &0.4410  &\textbf{0.5608} &0.9695  &0.8753 &\textbf{0.8104} &0.9487 &0.7965 &0.6999 & 0.9165\\					
			\hline
			\multirow{1}{*}{ERFNet} 
			&2.1M &\textbf{0.9669}    &\textbf{0.6919}   &\textbf{0.3002}  &\textbf{0.4460}  &0.5585 &\textbf{0.9701}  &\textbf{0.8815} &0.8092 &\textbf{0.9799} &\textbf{0.7976} &\textbf{0.7030} &\textbf{0.9178}\\
			\hline
		\end{tabular}
	}
\end{table*}

\begin{figure}[htbp]
	\centering	
	\begin{minipage}{\linewidth}
		\centerline{\includegraphics[width=\textwidth]{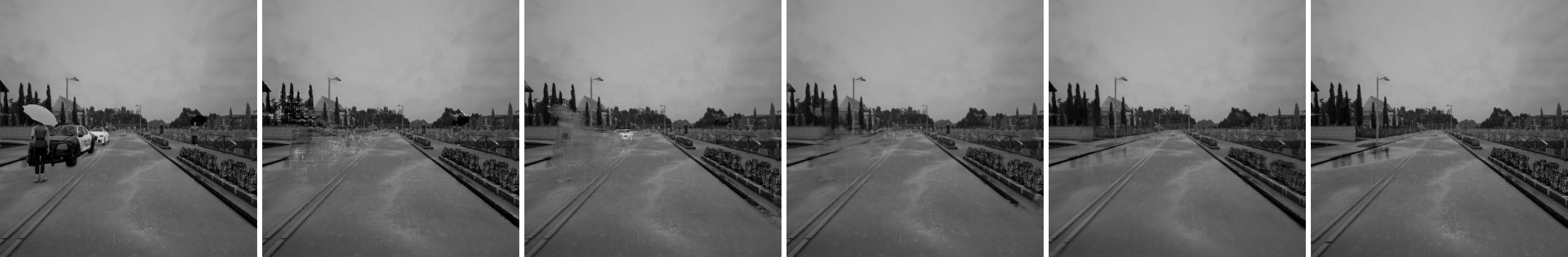}}
		\vspace{2pt}
	\end{minipage}
	\begin{minipage}{\linewidth}
	\centerline{\includegraphics[width=\textwidth]{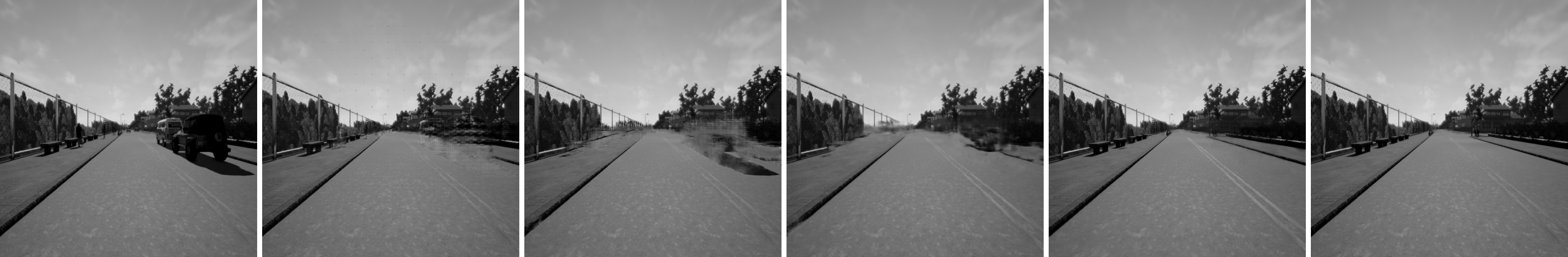}}
	\vspace{2pt}
	\end{minipage}
	\begin{minipage}{\linewidth}
	\centerline{\includegraphics[width=\textwidth]{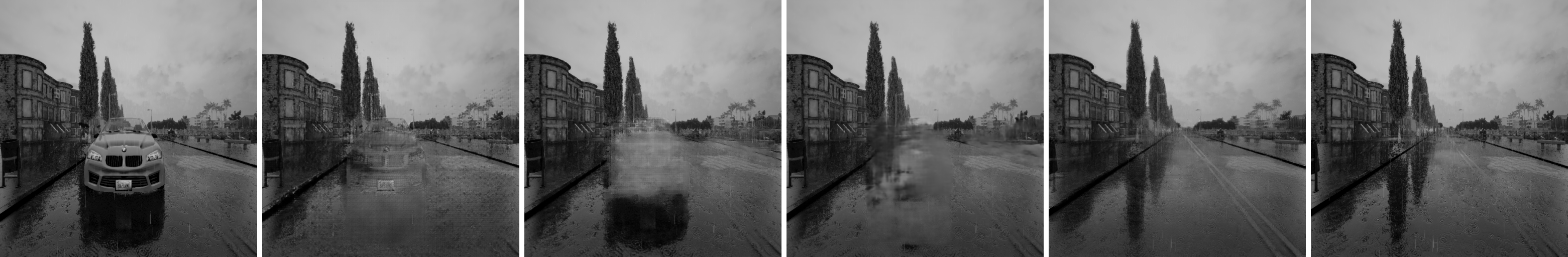}}
	\vspace{2pt}
	\end{minipage}
	\begin{minipage}{0.15\linewidth}
		\centerline{\footnotesize{(a) Dynamic}}
	\end{minipage}
	\begin{minipage}{0.15\linewidth}
		\centerline{\footnotesize{(b) Pix2Pix}}
	\end{minipage}
	\begin{minipage}{0.15\linewidth}
	\centerline{\footnotesize{(c) MGAN}}
	\end{minipage}
	\begin{minipage}{0.18\linewidth}
	\centerline{\footnotesize{(d) SRMGAN}}
	\end{minipage}
	\begin{minipage}{0.15\linewidth}
		\centerline{\footnotesize{(e) SSGGNet}}
	\end{minipage}
	\begin{minipage}{0.15\linewidth}
		\centerline{\footnotesize{(f) GT}}
	\end{minipage}
	\caption{Visual evaluations on dynamics-to-static image translation.}
	\label{fig_img}
\end{figure}

\textbf{Image Translation.} Pix2Pix~\cite{isola2017image}, MGAN~\cite{bescos2019empty}, and SRMGAN~\cite{2020Empty} are considered for performance evaluation and comparison.  As shown in Fig.~\ref{fig_img}, SSGGNet is capable of eliminating dynamic objects as well as their shadows and generating visually realistic static images with high semantic consistency, although slight blur is noticed. In contrast, without the guidances of static semantics, Pix2Pix, MGAN and SRMGAN produce blur and artifacts within dynamic regions. Numerical performance comparisons in Table~\ref{tab_img} further demonstrate that the reconstruction of static images greatly benefits from the guidances of static semantics. Besides, we consider a variant of SSGGNet by utilizing the estimated segmentation map $M$ instead of the semantic probability map $S$ as its inputs, denoted as SSGGNet-SM. As shown in Table~\ref{tab_img},  this change leads to obvious performance degradation. This can be explained by the fact that using semantic probability map as inputs helps SSGGNet reduce noises from the segmentation network.

\begin{table}[htb]
	\centering
	\caption{Evaluations on image translation. $\uparrow$ indicates Higher is better while $\downarrow$ indicates Lower is better.}
	\label{tab_img}
	\resizebox{0.42\textwidth}{!}{
		\begin{tabular}{lcccc}
			\hline
			Method\hspace{0.5cm} & L1(\%)$\downarrow$ & L2(\%)$\downarrow$ & PSNR$\uparrow$ & SSIM$\uparrow$ \\
			\hline 
			Pix2Pix&3.220     &0.451   &24.483   &0.744 \\
			MGAN&2.903 &0.466 &24.595 &0.783 \\
			SRMGAN& 2.991 &0.414 &25.011 &\textbf{0.799} \\
			
			SSGGNet&\textbf{2.452} &\textbf{0.327}   &\textbf{26.316}   &0.793 \\
			SSGGNet-SM& 2.763 &0.391 &25.363 &0.773 \\
			
			\hline
		\end{tabular}%
	}
\end{table}%

\subsection{Evaluations on Multi-modal VPR}
\begin{figure}[htb]
	\centering
	\includegraphics[height=0.19\textheight, width=0.9\linewidth]{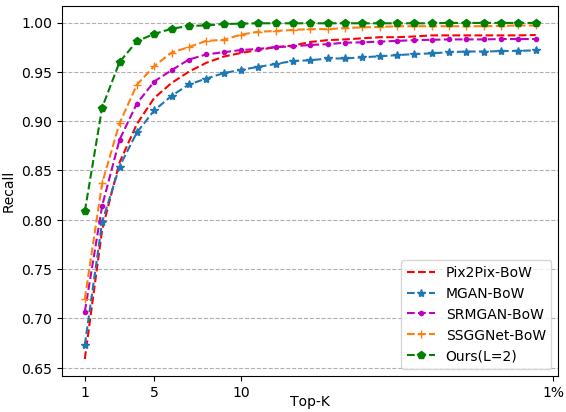}
	\caption{Recall curves of different methods on VPR.}
	\label{fig_recall}
\end{figure}

We consider Pix2Pix-BoW, MGAN-BoW, SRMGAN-BoW, and SSGGNet-BoW for performance comparison. These four baselines rely on BoW features extracted from the recovered static images from Pix2Pix, MGAN, SRMGAN and SSGGNet, respectively, to measure image similarity. In contrast, our approach exploits both BoW and SPM features. Recall accuracy of different models are presented in Table~\ref{tab_recall}. We notice that our method performs best and improves the second-performing SSGGNet-Bow by a large margin. This observation suggests the importance of SPM-based semantic features. Besides, we observe that SSGGNet-BoW outperforms all of Pix2Pix-BoW, MGAN-Bow and SRMGAN-Bow, which further validates the effectiveness of utilizing static semantics to guide the generation of static images. The recall curves of all methods are shown in Fig~\ref{fig_recall}. Note that our semantic segmentation and image translation procedures cost about  20ms and 15ms, respectively, when the resolution of query images is $512\times512$ pixels; and encoding two modalities and performing retrieval require about 22ms. Therefore, our entire VPR pipeline runs over 18fps in near real-time.

\vspace{-0.2cm}

\vspace{-0.2cm}
\begin{table}[htb]
	\centering
	\caption{Evaluations on recall accuracy.}
	\label{tab_recall}
	\resizebox{0.44\textwidth}{!}{
		\begin{tabular}{lcccc}
			\hline
			Method\hspace{0.5cm}   &R@1                     &R@5                &R@10                 &R@1\%\\
			\hline 
			Pix2Pix-BoW                  &0.6591                   &0.9234              &0.9691                &0.9873 \\
		        MGAN-BoW                    &0.6728                  &0.9108              &0.9520                 &0.9717  \\
		        SRMGAN-BoW               &0.7068                  &0.9402              &0.9721                 &0.9833 \\
		        SSGGNet-BoW               &0.7198                  &0.9557              &0.9877                 &0.9970 \\
			Ours(L=2)                       &\textbf{0.8086}      &\textbf{0.9884}   &\textbf{0.9988}    &\textbf{0.9996} \\
			\hline
		\end{tabular}
	}
\end{table}


\subsection{Ablation Study}

To investigate the influence of the number of pyramid levels on VPR, we report the recall accuracy and coding time of our approach under varying $L$ in Table~\ref{tab_ablation_L}. Here, coding time refers to the execution time on CPU to finish the multi-modal coding for each query. We observe that an increased $L$ brings improvements in recall accuracy. This could be explained by the more detailed description of the spatial distribution of semantic features. However, as $L$ increases, the growth rate of recall accuracy decreases, and more time is required to process each query. Based on the table, we could see that $L=2$ achieves a better tradeoff between accuracy and speed.
\vspace{-0.8cm}
\begin{table}[htb]
	\centering
	\caption{Ablation study on pyramid level $L$.}
	\label{tab_ablation_L}
	\resizebox{0.48\textwidth}{!}{
		\begin{tabular}{lccccc}
			\hline
			Method & R@1 & R@5  &  R@10  & R@1\% & Coding Time\\
			\hline 
			Ours(L=0)&0.7217 &0.9588   &0.9911   &0.9993 &15.65 ms\\
			Ours(L=2)&0.8086 &0.9884   &0.9988   &0.9996 &22.05 ms \\
			Ours(L=4)&0.8146 &0.9889   &0.9996   &1.0000 &34.92 ms \\
			Ours(L=6)&0.8198 &0.9911   &1.0000   &1.0000 &172.96 ms\\
			\hline
		\end{tabular}
	}
\end{table}

\section{Conclusion}
In this letter, we explore multi-modal features to enhance visual place recognition in dynamic environments. A novel multi-modal dynamics-invariant perception network is proposed to reconstruct static semantics and static image from dynamic frame.  Furthermore, we introduce a novel coding strategy to exploit semantic-visual cues for landmark retrieval. Experimental results prove that the generation of static images greatly benefits from the guidances of static semantics, and incorporating semantic features greatly improves the accuracy of place recognition.  In future, we will extend our work to the more challenging task where ground-truth static images are taken under different illuminations. Last but not least, we hope to extend our method to long-term visual localization \cite{2020Learning}, semantic SLAM \cite{2016SemanticSLAM}, semantic mapping \cite{2020RGB}, etc.

\balance
\bibliographystyle{IEEEtran}
\bibliography{IEEEabrv,mynewlib}

\end{document}